\documentstyle[12pt,mkstyle]{article}

\title{Independence, Conditionality and
Structure of
 Dempster-Shafer Belief Functions}
\shorttitle{Structure of Belief Functions}

\newcommand{\AbbEins}
{
\begin{figure}
a) \begin{picture}(4,4.0)
\put(1,2){\circle*{0.2}} 
\put(0.7,2.3){$A$} 
\put(2,3){\circle*{0.2}} 
\put(2.3,3.3){$B$} 
\put(3,2){\circle*{0.2}} 
\put(3.3,2.3){$C$} 
\put(3,1){\circle*{0.2}} 
\put(3.3,0.7){$D$} 
\put(1,1){\circle*{0.2}} 
\put(0.7,0.7){$E$} 
\put(0.5,1.6){\line(1,0){3}}
\put(0.5,3.6){\line(1,0){3}}
\put(0.5,1.6){\line(0,1){2}}
\put(3.5,1.6){\line(0,1){2}}
\put(2.7,0.3){\line(1,0){1}}
\put(2.7,0.3){\line(0,1){2}}
\put(3.7,2.3){\line(-1,0){1}}
\put(3.7,2.3){\line(0,-1){2}}
\put(0.7,0.7){\line(1,0){2.6}}
\put(0.7,0.7){\line(0,1){0.6}}
\put(3.3,1.3){\line(-1,0){2.6}}
\put(3.3,1.3){\line(0,-1){0.6}}
\put(0.3,0.3){\line(1,0){1}}
\put(0.3,0.3){\line(0,1){2}}
\put(1.3,2.3){\line(-1,0){1}}
\put(1.3,2.3){\line(0,-1){2}}

\end{picture}
b) 
\begin{picture}(5,5.0)
\put(1,2){\circle*{0.2}} 
\put(0.7,2.3){$A$} 
\put(2,3){\circle*{0.2}} 
\put(2.3,3.3){$B$} 
\put(3,2){\circle*{0.2}} 
\put(3.3,2.3){$C$} 
\put(4.0,1){\circle*{0.2}} 
\put(4.3,0.7){$D$} 
\put(4,3){\circle*{0.2}} 
\put(4.3,2.7){$F$} 
\put(1,1){\circle*{0.2}} 
\put(0.7,0.7){$E$} 
\put(0.5,1.6){\line(1,0){3}}
\put(0.5,3.6){\line(1,0){3}}
\put(0.5,1.6){\line(0,1){2}}
\put(3.5,1.6){\line(0,1){2}}
\put(1.7,2.7){\line(1,0){2.6}}
\put(1.7,2.7){\line(0,1){0.6}}
\put(4.3,3.3){\line(-1,0){2.6}}
\put(4.3,3.3){\line(0,-1){0.6}}
\put(3.8,0.1){\line(0,1){3.8}}
\put(3.8,0.1){\line(1,0){0.8}}
\put(4.6,3.9){\line(0,-1){3.8}}
\put(4.6,3.9){\line(-1,0){0.8}}
\put(2.7,0.3){\line(1,0){1.5}}
\put(2.7,0.3){\line(0,1){2}}
\put(4.2,2.3){\line(-1,0){1.5}}
\put(4.2,2.3){\line(0,-1){2}}
\put(0.7,0.7){\line(1,0){3.6}}
\put(0.7,0.7){\line(0,1){0.6}}
\put(4.3,1.3){\line(-1,0){3.6}}
\put(4.3,1.3){\line(0,-1){0.6}}
\put(0.3,0.3){\line(1,0){1}}
\put(0.3,0.3){\line(0,1){2}}
\put(1.3,2.3){\line(-1,0){1}}
\put(1.3,2.3){\line(0,-1){2}}

\end{picture}
\caption{An example of hypergraphs a) with b) without compatible belief 
network}  \label{abbeins}
\end{figure}
} 

\newcommand{\AbbZwei}
{
\begin{figure}
a)
\begin{picture}(4,4.0)
\put(1,2){\circle*{0.2}} \put(0.7,2.3){$A$} 
\put(2,3){\circle*{0.2}} \put(2.3,3.3){$B$} 
\put(3,2){\circle*{0.2}} \put(3.3,2.3){$C$} 
\put(3,1){\circle*{0.2}} \put(3.3,0.7){$D$} 
\put(1,1){\circle*{0.2}} \put(0.7,0.7){$E$} 
\put(1,2){\vector(1,1){0.9}}
\put(3,2){\vector(-1,1){0.9}}
\put(3,2){\vector(0,-1){0.8}}
\put(3,1){\vector(-1,0){1.8}}
\put(1,1){\vector(0, 1){0.8}}

\end{picture}
b)
\begin{picture}(4,4.0)
\put(1,2){\circle*{0.2}} \put(0.7,2.3){$A$} 
\put(2,3){\circle*{0.2}} \put(2.3,3.3){$B$} 
\put(3,2){\circle*{0.2}} \put(3.3,2.3){$C$} 
\put(3,1){\circle*{0.2}} \put(3.3,0.7){$D$} 
\put(1,1){\circle*{0.2}} \put(0.7,0.7){$E$} 
\put(1,2){\vector(1,1){0.9}}
\put(3,2){\vector(-1,1){0.9}}
\put(3,1){\vector(0, 1){0.8}}
\put(3,1){\vector(-1,0){1.8}}
\put(1,1){\vector(0, 1){0.8}}

\end{picture}
c)
\begin{picture}(4,4.0)
\put(1,2){\circle*{0.2}} \put(0.7,2.3){$A$} 
\put(2,3){\circle*{0.2}} \put(2.3,3.3){$B$} 
\put(3,2){\circle*{0.2}} \put(3.3,2.3){$C$} 
\put(3,1){\circle*{0.2}} \put(3.3,0.7){$D$} 
\put(1,1){\circle*{0.2}} \put(0.7,0.7){$E$} 
\put(1,2){\vector(1,1){0.9}}
\put(3,2){\vector(-1,1){0.9}}
\put(3,1){\vector(0, 1){0.8}}
\put(1,1){\vector( 1,0){1.8}}
\put(1,1){\vector(0, 1){0.8}}

\end{picture}
d)
\begin{picture}(4,4.0)
\put(1,2){\circle*{0.2}} \put(0.7,2.3){$A$} 
\put(2,3){\circle*{0.2}} \put(2.3,3.3){$B$} 
\put(3,2){\circle*{0.2}} \put(3.3,2.3){$C$} 
\put(3,1){\circle*{0.2}} \put(3.3,0.7){$D$} 
\put(1,1){\circle*{0.2}} \put(0.7,0.7){$E$} 
\put(1,2){\vector(1,1){0.9}}
\put(3,2){\vector(-1,1){0.9}}
\put(3,1){\vector(0, 1){0.8}}
\put(1,1){\vector( 1,0){1.8}}
\put(1,2){\vector(0,-1){0.8}}

\end{picture}
\caption{An example of belief networks corresponding to a hypergraph from 
Fig. 1.a)}  \label{abbzwei}
\end{figure}
} 

\newcommand{\AbbDrei}
{
\begin{figure}
a) \begin{picture}(6,6.0)
\put(1,2){\circle*{0.2}} 
\put(0.7,1.7){$Y_1$} 
\put(3,2){\circle*{0.2}} 
\put(2.7,1.7){$Y_2$} 
\put(5,2){\circle*{0.2}} 
\put(4.7,1.7){$Y_3$} 
\put(2,4){\circle*{0.2}} 
\put(1.7,4.3){$X_1$} 
\put(4,4){\circle*{0.2}} 
\put(4.3,4.3){$X_2$} 
\put(0.7,1.7){\line(0,1){3.1}}
\put(0.7,1.7){\line(1,0){4.7}}
\put(5.4,4.8){\line(0,-1){3.1}}
\put(5.4,4.8){\line(-1,0){4.7}}
\put(0.8,3.2){twig}
\put(0.1,0.1){\line(0,1){2.8}}
\put(0.1,0.1){\line(1,0){5.8}}
\put(5.9,2.9){\line(0,-1){2.8}}
\put(5.9,2.9){\line(-1,0){5.8}}
\put(0.3,0.3){branch}
\put(3.5,1.0){\circle{0.2}} 
\put(5.5,1.3){\circle{0.2}} 
\end{picture}
b) 
\begin{picture}(6,6.0)
\put(1,2){\circle*{0.2}} 
\put(0.7,1.7){$Y_1$} 
\put(3,2){\circle*{0.2}} 
\put(2.7,1.7){$Y_2$} 
\put(5,2){\circle*{0.2}} 
\put(4.7,1.7){$Y_3$} 
\put(2,4){\circle*{0.2}} 
\put(1.7,4.3){$X_1$} 
\put(4,4){\circle*{0.2}} 
\put(4.3,4.3){$X_2$} 
\put(2,4){\vector(1,0){1.8}} 
\put(1,2){\vector( 1,2){0.9}} 
\put(3,2){\vector(-1,2){0.9}} 
\put(5,2){\vector(-3,2){2.8}} 
\put(1,2){\vector( 3,2){2.8}} 
\put(3,2){\vector( 1,2){0.9}} 
\put(5,2){\vector(-1,2){0.9}} 
\end{picture}
\caption{
An example of a) a twig in hypertree and b) its fragment of belief 
network} \label{abbdrei} 
\end{figure}
} 

\newcommand{\AbbVierPic}[3]
{\begin{picture}(9,6.0)
\put(1,2){\circle*{0.2}} 
\put(0.7,1.7){$Y_1$} 
\put(3,2){\circle*{0.2}} 
\put(2.7,1.7){$Y_2$} 
\put(5,2){\circle*{0.2}} 
\put(4.7,1.7){$Y_3$} 
\put(2,4){\circle*{0.2}} 
\put(1.7,4.3){$X_1$} 
\put(4,4){\circle*{0.2}} 
\put(4.3,4.3){$X_2$} 
\put(0.7,1.7){\line(0,1){3.1}}
\put(0.7,1.7){\line(1,0){4.7}}
\put(5.4,4.8){\line(0,-1){3.1}}
\put(5.4,4.8){\line(-1,0){4.7}}
                   \put(0.8,3.2){twig: #1}
\put(0.1,0.1){\line(0,1){2.8}}
\put(0.1,0.1){\line(1,0){5.8}}
\put(5.9,2.9){\line(0,-1){2.8}}
\put(5.9,2.9){\line(-1,0){5.8}}
                  \put(0.3,0.3){branch: #2}
\put(3.5,1.0){\circle{0.2}} 
\put(5.5,1.3){\circle{0.2}} 
\put(4.0,0.7){\line(0,1){1.6}}
\put(4.0,0.7){\line(1,0){2.8}}
\put(6.8,2.3){\line(0,-1){1.6}}
\put(6.8,2.3){\line(-1,0){2.8}}
         \put(4.8,2.5){other hyperedge: #3}
\put(6.0,1.0){\circle{0.2}} 
\put(6.5,2.0){\circle{0.2}} 
\end{picture}

}
\newcommand{\AbbVier}
{
\begin{figure}
a) \AbbVierPic{$Bel_t$}{$Bel_b$}{$Bel_o$}

b) \AbbVierPic{$Bel_t$
}{$(Bel_b\ominus Bel_b ^{\downarrow t \cap b})\oplus Bel_b ^{\downarrow t 
\cap b}$ 
}{$(Bel_o\ominus Bel_o ^{\downarrow t \cap b \cap o})\oplus Bel_o 
^{\downarrow t \cap b \cap o}$ 
}

c) \AbbVierPic{$Bel'_t=Bel_t\oplus Bel_b ^{\downarrow t \cap b}\oplus Bel_o 
^{\downarrow t \cap b \cap o} $
}{$Bel_b\ominus Bel_b ^{\downarrow t \cap b}$ 
}{$Bel_o\ominus Bel_o ^{\downarrow t \cap b \cap o}$ 
} 
\end{figure}

\begin{figure}
d) \AbbVierPic{${Bel'}_t ^{| t \cap b}\oplus {Bel'}_t
^{\downarrow t \cap b} $
}{$Bel_b\ominus Bel_b ^{\downarrow t \cap b}$ 
}{$Bel_o\ominus Bel_o ^{\downarrow t \cap b \cap o}$ 
}

e) \AbbVierPic{${Bel'}_t ^{| t \cap b}$
}{$(Bel_b\ominus Bel_b ^{\downarrow t \cap b})\oplus {Bel'}_t
^{\downarrow t \cap b} $ 
}{$Bel_o\ominus Bel_o ^{\downarrow t \cap b \cap o}$ 
} 

\caption{An example of valuation transformation}  
\label{abbvier} 
\end{figure}
} 

\newcommand{\V}{{\bf V }}
\newcommand{\smcond}{{: \atop :}}
\newcommand{\oand}{\bigcirc\hspace*{-2mm}\land}
\newcommand{\Bem}[1]{}

\begin{document}
\machetitel
\begin{abstract}
Several approaches of structuring (factorization, decomposition) of
Dempster-Shafer joint belief functions from literature are reviewed with
special emphasis on their capability to capture independence from the point of
view of the claim that belief functions generalize bayes notion
of probability. It is demonstrated that Zhu and Lee's \cite{Zhu:93} logical
networks and Smets' \cite{Smets:93} directed acyclic graphs are unable to
capture statistical dependence/independence of bayesian networks
\cite{Pearl:88}. On the other hand, though Shenoy and Shafer's hypergraphs can
explicitly represent bayesian network factorization of bayesian belief
functions, they disclaim any need for representation of independence of
variables in belief functions. Cano et al. \cite{Cano:93} reject the
hypergraph representation of Shenoy and Shafer just on  grounds of missing
 representation of variable independence, but in their frameworks some
belief functions factorizable in Shenoy/Shafer  framework  cannot 
be factored.
The approach in \cite{Klopotek:93f} on the other hand combines the merits of
both Cano et al. and of Shenoy/Shafer approach in that for Shenoy/Shafer
approach no simpler factorization than that in \cite{Klopotek:93f} approach
exists and on the other hand all independences among variables captured in
Cano et al. framework and many more are captured in \cite{Klopotek:93f}
approach.
\end{abstract}
\section{Introduction}

The Dempster-Shafer Theory  or the Mathematical Theory of Evidence (MTE) 
\cite{Shafer:76}, 
\cite{Dempster:67} 
 shows one of possible ways of application of mathematical probability for
subjective evaluation and is intended to be a generalization of bayesian
theory of subjective probability \cite{Shafer:90ijar}.
Belief functions are deemed to generalize (finite discrete) probability
functions in that belief functions assign basic belief  mass 
to (non-empty) subsets of set of elementary events, whereas probability
functions  assign basic belief  mass only to elementary events.
It is frequently claimed that, though they comprise something more than just
probabilistic uncertainty, MTE belief function behavior reduces to
behavior of probability if probabilities are available \cite{Smets:93}. 
That is if a belief function assigns non-zero basic belief  mass only 
to  subsets of cardinality 1  of the set of elementary events, then it is
called bayesian belief function and considered as equivalent to probability
function. 

A known method of representation of joint probability distribution in (many)
discrete variables are so-called bayesian networks (as described e.g. in
\cite{Pearl:88}, \cite{Geiger:90}). The joint probability distribution 
$Pr(x_1,...,x_n)$ in variables $X_1, X_2, ..., X_n$, where for a node $X_i$
only variables with indices from the set $\pi(i)$ directly influence
the value of $X_i$, is expressed as:
$$Pr(x_1,...,x_n) = \prod_{i=1}^{n} Pr(x_i | x_{\pi (i)})$$

\noindent It is assumed that if we form a directed graph with nodes
representing variables and directed edges are (all and only) of the form
$(X_k->X_i)$ with $k\in\pi(i)$ then this graph is acyclic.
As a direct representation 
of a joint probability distribution in 
 15 discrete 
variables, with a domain with cardinality four each, would require
more than 1~000~000~000 storage cells, bayesian network representation 
will immensely contribute to reduction of storage requirement
if every variable is directly influenced by only a few other. Beside this,
however, they are known to represent qualitatively many (conditional)
independences among variables as well as to capture (a part of) causal
relations among variables.

Structuring is much more urgently needed for belief functions.   If we have a
MTE belief  distribution in 3 discrete variables, each with a domain of
cardinality 4, then the joint belief  distribution will be non-zero for
possibly  $2^{4^3}-1>10~000~000~000~000~000~000$ points ! A much more
elaborated handling of structure of a joint belief distribution may be
needed, but let us restrict ourselves to the modest requirement to
structure at  least as good as bayesian networks do for probability.

Several concepts of structuring belief functions have been  proposed. Shenoy
and Shafer \cite{Shenoy:90} have proposed factorizations of belief functions
along hypergraphs. Smets \cite{Smets:93} and Cano at al. \cite{Cano:93}
proposed factorization of belief function along  directed acyclic graphs (Both
proposals are radically different, nonetheless). Zhu and Lee \cite{Zhu:93}
proposed amendment of a logical network connectives with specialized belief
functions. Still another type of belief network, based on directed acyclic
graphs, has been proposed in \cite{Klopotek:93f}.

Within this paper, let us look at these proposals from the point of view of
the benchmark established by bayesian networks. Especially we will ask whether
or not these proposals cover structuring of probability distributions into a
bayesian network and whether they are capable of expressing
dependence independence relations 
among variables, especially for special case of probability distribution. 

Basic definitions from the MTE are given in Appendix.

\section{Logical Networks of Zhu and Lee}

Zhu and Lee \cite{Zhu:93} propose representation of a  knowledge base 
(thus the joint belief)
as a set of rules (and facts) amended by MTE-styled truth probability
intervals. Though a name "logical network" is never used in their paper, it is
clearly an intention of the authors to have one as they consider
forward and backward propagation for each type of basic logical connector 
(negation, and, or, material implication).
For a rule $r:A \rightarrow B$ its probability interval would be
$[r_L, r_U]$. This implies immediately $m(\overline{A}\lor B)=r_L$, $m(A\land
\overline{B}
)=1-r_U$, $m(X)=r_U-r_L$, where X=\{true,false\} is the logical universe
considered in the paper. The authors claim the necessity of permanent search
for common frame of discernment for combined rules and facts to be major
disadvantage of general MTE framework. Therefore they seek a way around by
restricting themselves to logical values of rules and facts. They derive
formulas for forward and backward propagation of uncertainty as well as for
combination of evidence.

On page 347 they propose e.g. the following "Modus Ponens":\\

\begin{tabular}{ll}
$A\rightarrow B:$: & $[r_L,r_U]$\\
$A:$              & $[a_L,a_U]$\\
\hline
$B:$: & $[b_L,b_U]$\\
\end{tabular}

with \\
$$b_L = \min\{1,\max\{0,(r_L+a_U-1)/a_U\}\}$$
$$b_U = \min\{1,\max\{0,[r_U+a_U-a_L(r_L+a_U-1)/a_U-1]/(a_U-a_L)\}\}$$

One could feel impressed by the simplicity of this and other formulas if not
the way they are derived. On page 345 we find the table for joint belief
distribution of A and B (Table II, proposed conjunction procedure for
$A\land B$)):\\
\begin{tabular}{rllll}
\hline
&& \multicolumn{3}{c}{B}\\
\cline{3-5}
&
\multicolumn{1}{c}{A}
  &\{t\},$b_L$  & \{f\},$1-b_U$  & \{t,f\},$b_U-b_L$ \\
\hline
  &\{t\},$a_L$        & \{t\},$m_{11}$& \{f\},$m_{12}$& \{t\},$m_{13}$\\
A & \{f\},$1-a_U$     & \{f\},$m_{21}$& \{f\},$m_{22}$& \{f\},$m_{23}$\\
  & \{t,f\},$a_U-a_L$ & \{t\},$m_{31}$& \{f\},$m_{32}$& \{t,f\},$m_{33}$\\
\hline
\end{tabular}

Out of this table the basic belief assignment
for implication 
 is (presumably) derived as follows:
$$m_{A\rightarrow B}(\{true\})=r_L=m_{11}+m_{21}+m_{22}+m_{23}+m_{31}$$
$$m_{A\rightarrow B}(\{false\})=1-r_U=m_{12}+m_{13}+m_{32}$$ 
$$m_{A\rightarrow B}(\{true, false\})=r_U-r_L=m_{33}$$ 

Up to this point one can little complain
(beside e.g. typing error in \{f\}column header of B). But subsequently
authors assume independence(!) of A and B, just:
$m_{11}=a_Lb_L$, $m_{12}=a_L(1-b_U)$, $m_{13}=a_L(b_U-b_L)$, 
$m_{21}=(1-a_U)b_L$, $m_{22}=(1-a_U)(1-b_U)$, $m_{23}=(1-a_U)(b_U-b_L)$, 
$m_{31}=(a_U-a_L)b_L$, $m_{32}=(a_U-a_L)(1-b_U)$, $m_{33}=(a_U-a_L)(b_U-b_L)$,
\\
Hence 
$$m_{A\rightarrow B}(\{true\})=
a_Lb_L+
(1-a_U)b_L+(1-a_U)(1-b_U)+(1-a_U)(b_U-b_L)+  
(a_U-a_L)b_L=$$
$$= a_Lb_L+ a_L(b_U-b_L)+ (1-a_U)+  (a_U-a_L)b_L= $$
$$= a_Lb_L+ 1-a_U+  a_Ub_L-a_Lb_L= $$
$$= a_Lb_L+ 1-a_U+  a_Ub_L-a_Lb_L= $$
$$= 1-a_U+  a_Ub_L=r_L $$

In a similar way we obtain an expression for 
$$m_{A\rightarrow B}(\{true, false\})=(a_U-a_L)\cdot(b_U-b_L)=$$
$$= a_U\cdot b_U-a_U\cdot b_L- a_L\cdot b_U+a_L\cdot b_L=$$
$$=r_U-r_L$$ 

Hence $$r_U=a_U b_U-a_U b_L- a_L b_U+a_L b_L + 1-a_U+  a_Ub_L=$$
$$=a_U b_U- a_L b_U+a_L b_L + 1-a_U$$

These are the formulas for $[r_L,r_U$ interval of a rule R as presented on
page 346. Equations for $r_L, r_U$ are  solved to obtain $b_L$ and
$b_U$ as given at the beginning of this section. Let us assume that logical
formulas A and B have the following joint distribution of probability of
truth: 

\begin{tabular}{rllll}
\hline
&& \multicolumn{3}{c}{B}\\
\cline{3-5}
&
\multicolumn{1}{c}{A}
  &\{t\},$b_L=0.3$  & \{f\},$1-b_U=0.7$  & \{t,f\},$b_U-b_L=0$ \\
\hline
  &\{t\},$a_L=0.4$        & 0.1 & 0.3 & 0\\
A & \{f\},$1-a_U=0.6$     & 0.2 & 0.4 & 0\\
  & \{t,f\},$a_U-a_L=0$ & 0   & 0   & 0\\
\hline
\end{tabular}

Hence $a_L=a_U=0.4$ and $b_L=b_U=0.3$. So 
$r_L= 1-0.4+0.3\cdot 0.4=0.48$ and 
$r_U=0.4\cdot 0.3- 0.4\cdot 0.3+ 0.4\cdot 0.3 + 1-0.4=0.48$.

Let us assume we know from somewhere that A is true nearly for sure that is
that probability of truth of A is $a_L=0.999, a_U=1$. Then from Zhu and Lee
formulas follows: 

$$b_L = \min\{1,\max\{0,(0.48+1-1)/1\}\}=0.48$$
$$b_U = \min\{1,\max\{0,
[0.48+1-0.999(0.48+1-1)/1-1]/(1-0.999)\}\}=$$
$$= \min\{1,\max\{0,
[0.48+0.999\cdot 0.48]/(1-0.999)\}\}=0.48$$

But if we look at the data then is is clear that the 
probability of truth of B given truth of A  is 0.25 and given the narrow
uncertainty bound on truth of A the conditional (obtained by Jeffrey's rule)
will not exceed 0.26.

This example demonstrates in a clear way that the interpretation of DST
proposed in \cite{Zhu:93} in no way supports the generally expressed claim
that DST can capture bayesian reasoning as a special case. Also the source
of the bug is obvious. If one assumes a priori the independence of facts and
hypotheses (both in bayesian and DST sense) then one shall not wonder that the
rule of inference tells nothing meaningful about the relationship between
variables considered.

\section{Directed Acyclic Graphs of Smets}

In his paper \cite{Smets:93} Smets attempts to generalize the bayesian theorem
(being foundation of probability propagation in bayesian networks, e.g.
\cite{Pearl:88}) in such a way as to enable propagation of beliefs in a
directed networks. For this purpose he  introduces a special notion of
conditional beliefs (page 6)
$$bel(B \smcond A)=bel(B \cup \overline{A})-bel(\overline{A}) \quad \forall
B\subseteq \Omega$$
$\Omega$ - set of all elementary events.

On page 5 he states that $bel(\emptyset)=0$.
 Let x be a subset of the set X,
$\theta$ a subset of the set $\Theta$. Then on page 8 he states that
$pl_X(x\smcond \theta)=pl_{X\times\Theta}(cyl(x)\smcond cyl(\theta))$. 
On page 9 he writes that $pl(A)=bel(\Omega)-bel(\overline{A})$. Though not
explicitly stated, we expect that also 
$pl_X(A\smcond \theta)=bel_X(X \smcond
\theta)-bel_X(\overline{x\smcond\theta})$ should hold. 
   So 
 $$bel(\overline{B} \smcond A)
=bel(\overline{B} \cup \overline{A})-bel(\overline{A})$$            
 $$bel(\Omega \smcond A)=bel(\Omega \cup \overline{A})-bel(\overline{A})$$
Hence 
 $$bel(\Omega \smcond A)-bel(\overline{B} \smcond A)=
bel(\Omega \cup \overline{A})-bel(\overline{A})
-bel(\overline{B} \cup \overline{A})+bel(\overline{A})=$$ $$=
bel(\Omega )-bel(\overline{B} \cup \overline{A})=$$ $$=
pl(\overline{\overline{B} \cup \overline{A}})=pl(B\cap A)=pl(B\smcond A)
$$
Hence 
$$pl_X(x\smcond \theta)=pl_{X\times\Theta}(cyl(x)\smcond cyl(\theta))
= 
bel_{X\times\Theta}(cyl(X)\smcond cyl(\theta))-
bel_{X\times\Theta}(\overline{cyl(x)}\smcond cyl(\theta))$$

But then we easily derive that 
$bel_X(x\smcond \theta)=bel_{X\times\Theta}(cyl(x)\smcond cyl(\theta))$.

 He gives also the formula that given
two belief distributions $bel_1, bel_2$ and $bel_{12}=bel_1\oand bel_2$
($\oand$ is a version of $\oplus$ which is not normalized) we have 
$$m_{12}(A)=\sum_{B\subseteq\Omega} m_1(A\smcond B)m_2(B)$$.
On page 12 he defines that two variables X and Y are said to be independent
iff $$bel_X(A\smcond y)=bel_X(A\smcond y'), \forall A\subseteq X, \forall
 y, y' \in Y, y\ne y'$$
and $$bel_Y(B\smcond x)=bel_Y(B\smcond x'), \forall B\subseteq Y, \forall x,
x'\in X, x\ne x'$$
Furthermore, for the set (of contexts) $\Theta=\{\theta_i, i=1,\dots,n\}$ he
defines that when two observations are independent whatever the context
$\theta_i$, then they are called conditionally independent.\\

Then on page 16 he requires that there is a $bel_\Theta(.\smcond x,y)$ such
that  $$bel_\Theta(.\smcond x,y)= bel_\Theta(.\smcond x) \oand
bel_\Theta(.\smcond y)$$\\

Let us consider the consequences. 
Let X, and Y be sets $\{x_p,x_q\}$ and $\{y_p,y_q\}$ resp.
Let $bel_{X\times Y}$ be a bayesian belief
distribution, that is  $m_{X\times Y}(A)$  greater than zero for some $A
\subseteq X \times Y$ with card(A)=1 (singletons) and elsewhere equal zero.
\Bem{
Furthermore let  us assume that m's sum up to 1
and let us 
"statistical independence" of X and Y, that is masses are distributed as
follows (#p_x+q_x=p_y+q_y=1)
\begin{tabular}{rlll}
\hline
&& \multicolumn{2}{c}{Y}\\
\cline{3-4}
&
\multicolumn{1}{c}{X}
  &\{$y_p$\}  & \{$y_q$\}\\
\hline
X &\{$x_p$\}     & $p_x\cdot p_y$  & $p_x\cdot q_y$ \\
  &\{$x_q$\}     & $q_x\cdot p_y$  & $q_x\cdot q_y$  & \\
\hline
\end{tabular}

Under which conditions are X and Y "cognitively independent" ?
Smets requires that
 $bel_X(x_p\smcond y_p)=bel_X(x_p\smcond y_q)$, 
 $bel_X(x_q\smcond y_p)=bel_X(x_q\smcond y_q)$, 
 $bel_Y(y_p\smcond x_p)=bel_Y(y_p\smcond x_q)$, 
 $bel_Y(y_q\smcond x_p)=bel_Y(y_q\smcond x_q)$, 

Now $bel_X(x_p\smcond y_p)
=bel_{X\times\Theta}(cyl(x_p)\smcond cyl(y_p))
=bel_{X\times\Theta}(cyl(x_p) \cup \overline{cyl(y_p)})
   - bel_{X\times\Theta}(\overline{cyl(y_p)})
= (p_x\cdot p_y+p_x\cdot q_y+q_x\cdot q_y)-(p_x\cdot q_y+q_x\cdot q_y)
= p_x\cdot p_y$, 
and
 $bel_X(x_p\smcond y_q) =  p_x\cdot q_y$, 
 $bel_X(x_q\smcond y_p) =  q_x\cdot p_y$, 
 $bel_X(x_q\smcond y_q) =  q_x\cdot q_y$, 
 $bel_Y(y_p\smcond x_p) =  p_x\cdot p_y$, 
 $bel_Y(y_p\smcond x_q) =  q_x\cdot p_y$, 
 $bel_Y(y_q\smcond x_p) =  p_x\cdot q_y$, 
 $bel_Y(y_q\smcond x_q) =  q_x\cdot q_y$. 
But 
 $bel_X(x_p\smcond y_p)=bel_X(x_p\smcond y_q)$ implies $p_y=q_y$, 
 $bel_Y(y_p\smcond x_p)=bel_Y(y_p\smcond x_q)$ implies $p_x=q_x$. 
Hence if variables X and Y are statistically independent, they are cognitively
independent if $p_x=q_x=p_y=q_y=0.5$
}

Let be given the following distribution -  basic belief assignment

\begin{tabular}{rlll}
\hline
&& \multicolumn{2}{c}{Y}\\
\cline{3-4}
&
\multicolumn{1}{c}{X}
  &\{$y_p$\}  & \{$y_q$\}\\
\hline
X &\{$x_p$\}     & $m_{pp}$   & $m_{pq}$  \\
  &\{$x_q$\}     & $m_{qp}$   & $m_{qq}$  \\
\hline
\end{tabular}

Under which conditions are X and Y "cognitively independent" ?
Smets requires that
 $bel_X(x_p\smcond y_p)=bel_X(x_p\smcond y_q)$, 
 $bel_X(x_q\smcond y_p)=bel_X(x_q\smcond y_q)$, 
 $bel_Y(y_p\smcond x_p)=bel_Y(y_p\smcond x_q)$, 
 $bel_Y(y_q\smcond x_p)=bel_Y(y_q\smcond x_q)$, 

Following Smets' notation cyl(x) shall denote cylindric (vacuous) extension of
set x. Now $$bel_X(x_p\smcond y_p)
=bel_{X\times\Theta}(cyl(x_p)\smcond cyl(y_p))=$$ $$
=bel_{X\times\Theta}(cyl(x_p) \cup \overline{cyl(y_p)})
   - bel_{X\times\Theta}(\overline{cyl(y_p)})=$$ $$
= (m_{pp}+m_{pq}+m_{qq}) - (m_{pq}+m_{qq}) = m_{pp}$$, 

 $bel_X(x_p\smcond y_q) = m_{pq}$, 
 $bel_X(x_q\smcond y_p) = m_{qp}$, 
 $bel_X(x_q\smcond y_q) = m_{qq}$, 
 $bel_Y(y_p\smcond x_p) = m_{pp}$, 
 $bel_Y(y_p\smcond x_q) = m_{qp}$, 
 $bel_Y(y_q\smcond x_p) = m_{pq}$, 
 $bel_Y(y_q\smcond x_q) = m_{qq}$. 
But 
 $bel_X(x_p\smcond y_p)=bel_X(x_p\smcond y_q)$ implies $m_{pp}=m_{pq}$.
 $bel_X(x_q\smcond y_p)=bel_X(x_q\smcond y_q)$ implies $m_{qp}=m_{qq}$, 
 $bel_Y(y_p\smcond x_p)=bel_Y(y_p\smcond x_q)$ implies $m_{pp}=m_{qp}$.

Hence both are independent only if 

\begin{tabular}{rlll}
\hline
&& \multicolumn{2}{c}{Y}\\
\cline{3-4}
&
\multicolumn{1}{c}{X}
  &\{$y_p$\}  & \{$y_q$\}\\
\hline
X &\{$x_p$\}     & $m$   & $m$  \\
  &\{$x_q$\}     & $m$   & $m$  \\
\hline
\end{tabular}

(m - a constant equal 1/4).

That is cognitive independence of Smets does not cover statistical
independence
for bayesian belief functions, but rather is a very special case of it (for
uniform distributions).

Let us consider now three variables $X,Y,\Theta$ for conditional independence
of X,Y on $\Theta$. Let X,Y have domains as above, let $\Theta=\{\theta_p,
\theta_q\}$. 
Let the joint belief distribution basic belief assignment be as follows: \\

\noindent
for $\theta_p$
\begin{tabular}{rlll}
\hline
&& \multicolumn{2}{c}{Y}\\
\cline{3-4}
&
\multicolumn{1}{c}{X}
  &\{$y_p$\}  & \{$y_q$\}\\
\hline
X &\{$x_p$\}     & $m_{ppp}$   & $m_{pqp}$  \\
  &\{$x_q$\}     & $m_{qpp}$   & $m_{qqp}$  \\
\hline
\end{tabular}
for $\theta_q$
\begin{tabular}{rlll}
\hline
&& \multicolumn{2}{c}{Y}\\
\cline{3-4}
&
\multicolumn{1}{c}{X}
  &\{$y_p$\}  & \{$y_q$\}\\
\hline
X &\{$x_p$\}     & $m_{ppq}$   & $m_{pqq}$  \\
  &\{$x_q$\}     & $m_{qpq}$   & $m_{qqq}$  \\
\hline
\end{tabular}

When may X and Y be conditionally independent given $\Theta$ ?
Smets requires that among others 
 $$m_\Theta(\theta_p\smcond x_p,y_p)= m_\Theta(\theta_p\smcond x_p) \cdot
m_\Theta(\theta_p\smcond y_p)$$
 $$m_\Theta(\theta_p\smcond x_p,y_q)= m_\Theta(\theta_p\smcond x_p) \cdot
m_\Theta(\theta_p\smcond y_p)$$
 $$m_\Theta(\theta_p\smcond x_q,y_p)= m_\Theta(\theta_p\smcond x_q) \cdot
m_\Theta(\theta_p\smcond y_p)$$
 $$m_\Theta(\theta_p\smcond x_q,y_q)= m_\Theta(\theta_p\smcond x_q) \cdot
m_\Theta(\theta_p\smcond y_q)$$
 $$m_\Theta(\theta_q\smcond x_p,y_p)= m_\Theta(\theta_q\smcond x_p) \cdot
m_\Theta(\theta_q\smcond y_p)$$
 $$m_\Theta(\theta_q\smcond x_p,y_q)= m_\Theta(\theta_q\smcond x_p) \cdot
m_\Theta(\theta_q\smcond y_q)$$
 $$m_\Theta(\theta_q\smcond x_p,y_p)= m_\Theta(\theta_q\smcond x_q) \cdot
m_\Theta(\theta_q\smcond y_p)$$
 $$m_\Theta(\theta_q\smcond x_p,y_q)= m_\Theta(\theta_q\smcond x_q) \cdot
m_\Theta(\theta_q\smcond y_q)$$

One can easily check that for bayesian belief functions $bel(B \cup
\overline{A})=bel(B \cap A) + bel(\overline{A})$. Then $bel(B \smcond A) =
bel(B \cap A)$. But $bel(B \cap A)$ is the sum of m-function values for all
singleton subsets of  $B \cap A$.  

 This actually means that: \\

 $$m_{\Theta\times X \times Y}(\theta_p,x_p,y_p)=
(m_{\Theta\times X \times Y}(\theta_p,x_p,y_p)+
m_{\Theta\times X \times Y}(\theta_p,x_p,y_q))
 \cdot$$ $$ \cdot
(m_{\Theta\times X \times Y}(\theta_p,x_p,y_p)+
m_{\Theta\times X \times Y}(\theta_p,x_q,y_p))$$
etc. 
Hence 
 $$m_{\Theta\times X \times Y}(\theta_p,x_p,y_p)=
m_{\Theta\times X \times Y}(\theta_p,x_p,y_p)^2+
m_{\Theta\times X \times Y}(\theta_p,x_p,y_q) \cdot 
m_{\Theta\times X \times Y}(\theta_p,x_p,y_p)+ $$ $$+      
m_{\Theta\times X \times Y}(\theta_p,x_p,y_p)+ \cdot 
m_{\Theta\times X \times Y}(\theta_p,x_q,y_p)
$$
etc.
Hence 
 $$m_{\Theta\times X \times Y}(\theta_p,x_p,y_p) 
 + m_{\Theta\times X \times Y}(\theta_p,x_p,y_q) 
 + m_{\Theta\times X \times Y}(\theta_p,x_q,y_p) 
 + m_{\Theta\times X \times Y}(\theta_p,x_q,y_q) 
=$$ $$ =     
m_{\Theta\times X \times Y}(\theta_p,x_p,y_p)^2+
m_{\Theta\times X \times Y}(\theta_p,x_p,y_p) \cdot 
m_{\Theta\times X \times Y}(\theta_p,x_p,y_q)+$$ $$+     
m_{\Theta\times X \times Y}(\theta_p,x_p,y_p)+ \cdot 
m_{\Theta\times X \times Y}(\theta_p,x_q,y_p)
+$$ $$+     
m_{\Theta\times X \times Y}(\theta_p,x_p,y_q)^2+
m_{\Theta\times X \times Y}(\theta_p,x_p,y_q) \cdot 
m_{\Theta\times X \times Y}(\theta_p,x_p,y_p)+$$ $$+     
m_{\Theta\times X \times Y}(\theta_p,x_p,y_q)+ \cdot 
m_{\Theta\times X \times Y}(\theta_p,x_q,y_q)
+$$ $$+     
m_{\Theta\times X \times Y}(\theta_p,x_q,y_p)^2+
m_{\Theta\times X \times Y}(\theta_p,x_q,y_p) \cdot 
m_{\Theta\times X \times Y}(\theta_p,x_p,y_q)+$$ $$+     
m_{\Theta\times X \times Y}(\theta_p,x_q,y_p)+ \cdot 
m_{\Theta\times X \times Y}(\theta_p,x_p,y_p)
+$$ $$+     
m_{\Theta\times X \times Y}(\theta_p,x_q,y_q)^2+
m_{\Theta\times X \times Y}(\theta_p,x_q,y_q) \cdot 
m_{\Theta\times X \times Y}(\theta_p,x_p,y_p)+$$ $$+     
m_{\Theta\times X \times Y}(\theta_p,x_q,y_q)+ \cdot 
m_{\Theta\times X \times Y}(\theta_p,x_p,y_q)
$$

Hence 
 $$m_{\Theta\times X \times Y}(\theta_p,x_p,y_p) 
 + m_{\Theta\times X \times Y}(\theta_p,x_p,y_q) +$$ $$
 + m_{\Theta\times X \times Y}(\theta_p,x_q,y_p) 
 + m_{\Theta\times X \times Y}(\theta_p,x_q,y_q) 
=$$ $$ =
(m_{\Theta\times X \times Y}(\theta_p,x_p,y_p) 
 + m_{\Theta\times X \times Y}(\theta_p,x_p,y_q) +$$ $$ 
 + m_{\Theta\times X \times Y}(\theta_p,x_q,y_p) 
 + m_{\Theta\times X \times Y}(\theta_p,x_q,y_q) )^2
$$

That is $bel_{\Theta\times X \times Y}(\theta_p)=
bel_{\Theta\times X \times Y}(\theta_p)^2$.
Shall we take seriously the assumption from page  that
$\sum_{A\subseteq\Omega}m(A)=1$, then either 
 $bel_{\Theta\times X \times Y}(\theta_p)=1$  and 
 $bel_{\Theta\times X \times Y}(\theta_q)=0$  or  
 $bel_{\Theta\times X \times Y}(\theta_p)=0$  and 
 $bel_{\Theta\times X \times Y}(\theta_q)=1$.    
But this actually means that variable $\Theta$ does not influence the joint
distribution of X,Y at all. Just the notion of conditional independence for
bayesian belief networks of Smets is devoid of any meaning. 

\section{Hypergraphs of Shenoy and Shafer}

In \cite{Shenoy:90}, Shenoy and Shafer proposed a general framework
for uncertainty propagation if uncertainty is structured along a hypergraph.
Their framework covers both probability and belief functions. 

{\em Hypergraphs}: A nonempty set H of nonempty subsets of a finite set S be 
called 
a hypergraph on S. The elements of H be called hyperedges. Elements of S be 
called vertices. H and H' be both hypergraphs on S, then we call a 
hypergraph H' a {\em reduced hypergraph} of the 
hypergraph H, iff for every $h'\in H'$ also  $h'\in H$ holds, and for 
every  $h \in H$ there exists such a $h' \in H'$ that $h \subseteq h'$.
A hypergraph H {\em covers} a hypergraph H' iff for every $h'\in H'$ there 
exists such a $h\in H$ that $h'\subseteq h$.

{\em Hypertrees}: t and b be distinct hyperedges in a hypergraph H, $t \cap 
b\neq 
\emptyset$, and b contains every vertex of t that is contained in a hyperedge 
of H other than t; if $X\in t$ and $X\in h$, where $h\in H$ and $h\neq t$, 
then $X\in b$. Then we call t a twig of H, and we call b a branch for t. A 
twig may have more than one branch. 
We call a hypergraph a hypertree if there is an ordering of its hyperedges, 
say $h_1,h_2,...,h_n$ such that $h_k$ is a twig in the hypergraph 
$\{h_1,_h2,...,h_k\}$ whenever $2 \leq k \leq n$. We call  any  such 
ordering of 
hyperedges a hypertree construction sequence for the hypertree. The first 
hyperedge in the hypertree construction sequence be called the root of the 
hypertree construction sequence. 

Please refer to the paper of Shenoy and Shafer \cite{Shenoy:90} on notions of
Markov trees,  
variables ({\bf V}), valuations (VV), valuations on a set of variables h 
($VV_h$), and proper valuations, combination 
operator $\oplus: VV \times VV \rightarrow VV$, marginalization operator 
 $\downarrow h:
\bigcup \{ VV_g| g \subseteq h\} \rightarrow VV_h$, the  axiomatic 
framework 
and the local computation method of Shenoy and Shafer. We recall here only the
definitions of: 
 
{\em Factorization}: Suppose A is a valuation on a finite set of variables \V,
and suppose HV is a hypergraph on \V. If A is equal to the combination of 
valuations of all hyperedges h  of HV then we say that A factors on HV.

The valuation for MTE is simply the belief function. 
For MTE the fact that  a belief
function Bel defined for the set of variables \V factors over a hypergraph
HV means that it may be represented
as
$$Bel = \bigoplus_{h;h \in HV} Bel ^{h}$$
where   $Bel ^{h}$ is a (different) belief function defined over the set of
variables h.

{\em Conditioning}: Suppose $Bel$ is a belief distribution,  and $Bel_E$ is 
an indicator potential   capturing the 
evidence E
(that is $Bel_E$ is such a belief function that $M_E(E)=1$ and for any A
different from E m(A)=0). Then conditional belief function conditioned on E,
 $Bel(.||E)$,
 is defined as  $Bel(.||E)=Bel \oplus Bel_E$.%

Actually the propagation of belief in the hypergraph via Shenoy/Shafer
propagation of uncertainty means
for a given hypothesis variable $V_i$ 
 calculation of $Bel(.||E) ^{\downarrow V_i}$
($\downarrow V_i$ means projection of belief function onto the subspace of the
single variable $V_i$) for the given
belief function Bel, the factorization of which along a hypergraph is known,
and for the evidence E  in such a way as to avoid the calculation of the
complete function $Bel$ and $Bell(.||E)$ as a in-between result - because  
both functions may consume too much memory. To manage it, the underlying
hypergraph is first transformed to a hypertree and thereafter the computations
are quick. \\
 
Shenoy and Shafer consider it unimportant whether or not the 
factors $Bel ^{h}$
factorization of $Bel$
should refer to any notion of conditionality. 
In fact,
one may easily construct a belief function factorization in which
no proper subset of  factors can  tell  anything  about  marginal 
distribution
of any variable belonging to the factors of this subset. \\

The hypergraph makes the impression of apparent grater generality than
bayesian networks of Pearl \cite{Pearl:88} because in case of bayesian belief
functions simpler factorization is possible (less factors, fewer variables)
than that along a bayesian network. However, in \cite{Klopotek:93f} it has
been shown that this is only a superficial effect because the real propagation
is run on hypertrees, and not in general type hypergraphs, and then the
generality of Shenoy \& Shafer factorization gives nothing beyond that of
bayesian network factorization (see also below).

\section{Cano's et al. A Priori Conditionals in Directed Acyclic Graphs}

Cano et al. in \cite{Cano:93} proposed a generalization of Pearl's bayesian
networks to capture DS belief distributions instead of Shenoy/Shafer
hypergraphs. They argue: "graphical structures used to represent
relationships among variables in our      work are     Pearl's causal networks
\cite{Pearl:88}, not Shenoy/Shafer's hypergraphs, because the former are more
appropriate to represent independence relationships among variables in a
direct way." (p.257). On page 262 (Definition 2) they define
a belief function $Bel$
 (a priori)
conditional belief function conditioned on variable set  $h$ 
by  requiring  $Bel ^{\downarrow h}$ to be a vacuous belief function. It is
easily checked that this notion of conditional belief functions allows to
represent statistically conditionally independent variables 
of a bayesian belief network as a priori  conditionally independent variables 
in Cano's et al. sense.

However, it cannot handle other belief functions which could be expressed 
in terms of a Dempster Rule of Combination.

As an example please verify, that the belief function $Bel_{12}$
$$Bel_{12}=Bel_1\oplus Bel_2$$

\noindent
with focal points for $Bel_1$, $Bel_2$ ($Bel_1$ defined for variables X,Y,
$Bel_2$ for variables X,Z, domains of variables: X: \{$x_1,x_2$\},  
Y: \{$y_1,y_2$\},  
Z: \{$z_1,z_2$\})
  
\begin{tabular}{ll}
set & $m_1(set)$\\
\hline
\{$(x_1,y_1), (x_1,y_2),$\\ $ (x_2,y_1), (x_2,y_2)$\}        &    0.1\\
\{$(x_1,y_1)$\}        &    0.2\\
\{$(x_1,y_2)$\}        &    0.25\\
\{$(x_2,y_1)$\}        &    0.3\\
\{$(x_2,y_2)$\}        &    0.15\\
\hline \quad\\
\end{tabular} 
\begin{tabular}{ll}
set & $m_2(set)$\\
\hline
\{$(x_1,z_1), (x_1,z_2),$\\$ (x_2,z_1), (x_2,z_2)$\}        &    0.2\\
\{$(x_1,z_1)$\}        &    0.2\\
\{$(x_1,z_2)$\}        &    0.3 \\
\{$(x_2,z_1)$\}        &    0.25\\
\{$(x_2,z_2)$\}        &    0.05\\
\hline \quad\\
\end{tabular} 

\noindent
cannot be represented in a structured manner as a product of a normal and
conditional belief function in sense of Cano et al. In this sense it is
immediately visible, that Shenoy-Shafer hypergraphs allow for more efficient
structuring of belief functions than Cano's et al. directed acyclic graph
representation. 

\section{Generalized belief networks}

The axiomatization system of Shenoy/Shafer refers to the notion of
factorization along a hypergraph. However, the actual propagation algorithm
operates on hypertrees. We investigate below implications
of this disagreement
%
%
\begin{df} \cite{Klopotek:93f}
We define               a mapping $\ominus: VV \times VV 
\rightarrow VV$ called decombination such that: 
if $Bel_{12}=Bel_1 \ominus Bel_2$ then $Bel_1=Bel_2 \oplus Bel_{12}$.
 \end{df}
In case of probabilities, decombination means memberwise division: 
$Pr_{12}(A)=Pr_1(A)/Pr_2(A)$. In case of DS pseudo-belief functions it means 
the operator $\ominus$ yielding a DS pseudo-belief function such that: 
whenever $Bel_{12}=Bel_1 \ominus Bel_2$ 
then $Q_{12}(A)=c \cdot Q_1/Q_2$. Both for probabilities and for DS belief 
functions decombination may be not uniquely determined. Moreover, for DS 
belief functions not always a decombined DS belief function will exist. Hence 
we extend the domain to DS pseudo-belief functions which is closed under this 
operator. We claim here without a proof (which is simple) that DS 
pseudo-belief 
functions fit the axiomatic framework of Shenoy/Shafer. Moreover, we claim 
that if an (ordinary) DS  belief  function  is  represented  by  a 
factorization in 
DS pseudo-belief functions, then any propagation of uncertainty yields the 
very 
same results as when it would have been factored into ordinary DS belief 
functions. 
\begin{df} \cite{Klopotek:93f}
By anti-conditioning $|$ of a belief function $Bel$ on a set of variables $h$ 
we understand the transformation: $Bel ^{|h}= Bel \ominus Bel ^{\downarrow 
h}$. 
\end{df}
Notably, anti-conditioning means in case of probability functions proper 
conditioning. Notice that due to the fact that $\ominus$ does not provide with
a unique result, so also anti-conditioning may yield many pseudo-belief
functions none of which is particularly distinguished. 
Let 
us define now the general notion of belief networks.:
\begin{df} \cite{Klopotek:93f}
 A 
belief 
 network is a pair (D,Bel) where D is a DAG (directed acyclic graph)
and Bel  is a belief 
distribution called the {\em underlying distribution}. Each node i in D 
corresponds to a variable $X_i$  in Bel, a set of nodes I corresponds to a 
set of variables $X_I$ and $x_i, x_I$
 denote values drawn from the domain of $X_i$ 
 and from the (cross product) domain of $X_I$ respectively. Each node in the 
network  is regarded as a storage cell for any  distribution 
$Bel ^{\downarrow \{X_i\} \cup X_{\pi (i)} |  X_{\pi (i)} }$
 where $X_{\pi (i)}$ is a set of nodes corresponding to 
the 
parent nodes $\pi(i)$ of i.  The underlying distribution represented by a 
 belief network is computed via:
$$Bel  = \bigoplus_{i=1}^{n}Bel ^{\downarrow \{X_i\} \cup X_{\pi (i)} |  
X_{\pi (i)} } $$
\end{df}
Please notice the local character of valuation of a node:
to valuate the node $i$ corresponding to variable $X_i$ only 
the marginal $Bel ^{\downarrow \{X_i\} \cup X_{\pi (i)}}$ needs to be known 
(e.g. from data) and not the entire belief distribution.

There exists a straight forward transformation of a belief network structure
into a hypergraph, and hence of 
a belief network into a hypergraph:
for every node i of the underlying DAG define a hyperedge as the set
$\{X_i\} \cup X_{\pi(i)}$; then the valuation of this hyperedge define as
$Bel ^{\downarrow \{X_i\} \cup X_{\pi(i)} | X_{\pi(i)}}$. We say that the 
hypergraph obtained in this way is {\em induced} by the belief network 
\AbbEins

Let us consider now the inverse operation: transformation of a valuated
hypergraph into a belief network.
As the first  stage we consider structures of a hypergraph and of a
belief network (the underlying DAG). we say that a belief network is 
{\em compatible} with a hypergraph 
iff the reduced set of hyperedges induced by    the belief network is 
identical with the reduced hypergraph. 

\begin{Bsp} 
Let us consider the following hypergraph (see Fig.\ref{abbeins}.a)):
\{\{A,B,C\}, \{C,D\}, \{D,E\}, \{A, E\}\}.
the following belief network structures are compatible with this hypergraph:
\{$A,C\rightarrow B$, $C\rightarrow D$, $D\rightarrow E$, $E\rightarrow A$\}
(see Fig.\ref{abbzwei}.a)),
\{$A,C\rightarrow B$, $D\rightarrow C$, $D\rightarrow E$, $E\rightarrow A$\},
(see Fig.\ref{abbzwei}.b)),
\{$A,C\rightarrow B$, $D\rightarrow C$, $E\rightarrow D$, $E\rightarrow A$\},
(see Fig.\ref{abbzwei}.c)),
\{$A,C\rightarrow B$, $D\rightarrow C$, $E\rightarrow D$, $A\rightarrow E$\}.
(see Fig.\ref{abbzwei}.d)),
\end{Bsp}

\AbbZwei

\begin{Bsp}
Let us consider the following hypergraph (see Fig.\ref{abbeins}.b)):
\{\{A,B,C\}, \{C,D\}, \{D,E\}, \{A, E\}, \{B,F\}, \{F,D\}\}.
No belief network structure is compatible with it.
\end{Bsp}
The missing compatibility is connected with the fact that
a hypergraph may represent a cyclic graph. 
Even if a compatible belief network has been found we may have troubles with 
 valuations. In Example 1 an unfriendly valuation of hyperedge 
\{A,C,B\} may require an edge AC in a belief network representing the same 
distribution, but  it will  make 
the 
hypergraph incompatible (as e.g. hyperedge \{A,C,E\} would be induced). This 
may be demonstrated as follows:
\begin{df}   \cite{Klopotek:93f}
If $X_J,X_K,X_L$ are three disjoint sets of variables of a distribution Bel, 
then $X_J,X_K$ are said to be conditionally independent given $X_L$ (denoted 
$I(X_J,X_K |X_L)_{Bel}$) iff 
 $$Bel ^{\downarrow X_J \cup X_K \cup X_L |  X_L} 
 \oplus  Bel ^{\downarrow   X_L } =
 Bel ^{\downarrow X_J  \cup X_L |  X_L} \oplus 
 Bel ^{\downarrow X_K \cup X_L |  X_L} 
 \oplus  Bel ^{\downarrow   X_L } $$
%
$I(X_J,X_K |X_L)_{Bel}$ is called a {\em 
conditional independence statement}
\end{df}
Let $I(J,K|L)_D$ denote d-separation in a graph \cite{Geiger:90}.:
\begin{th} \label{IDIBel}  \cite{Klopotek:93f}
Let $Bel_D=\{Bel|$(D,Bel) be a belief network\}. Then:\\
$I(J,K|L)_D$ iff $I(X_J,X_K |X_L)_{Bel}$ for all $Bel \in Bel_D$.
\end{th}
Now we see in the above example that nodes D and E d-separate nodes A and C. 
 Hence within any belief network based on one of the three DAGs mentioned A 
will 
be conditionally independent from C given D and E. But one can easily check 
that with general type of hypergraph valuation nodes A and C may be rendered 
dependent. 
 The sad result is, that 
\begin{th} \cite{Klopotek:93f}
Hypergraphs considered  by
 Shenoy/Shafer \cite{Shenoy:90} 
may for a given joint belief distribution have simpler structure than
(be properly covered by)
 the closest hypergraph induced by a 
belief network.
\end{th}
%
%
%
%
\AbbDrei
Notably, though the axiomatic system of Shenoy/Shafer refers to hypergraph
factorization of a joint belief distribution, the actual propagation is run 
on a hypertree (or more precisely, on one construction sequence of a 
hypertree, that is on Markov tree) covering that hypergraph. 
\Bem{
Covering a 
hypergraph with a hypertree is a trivial task, yet finding the optimal one 
(with 
as small number of variables in each hyperedge of the hypertree as possible) 
may be very difficult \cite{Shenoy:90}.}
 Let us look closer at the outcome of the process of covering with a reduced
 hypertree factorization, or more precisely, at the relationship of a 
hypertree construction 
sequence and a  belief network constructed out of it in the following way:
If $h_k$ is a twig in the sequence $\{h_1,...,h_k\}$ and $h_{i_k}$ its branch 
with $i_k<k$, then let us span the following directed edges in a belief 
network: First make a complete directed acyclic graph out of nodes 
$h_k-h_{i_k}$. Then add edges $Y_l \rightarrow X_j$ for every $Y_l \in 
h_k \cap h_{i_k}$ and every $X_j \in h_k-h_{i_k}$.  (see Fig.\ref{abbdrei}).
Repeat this for every k=2,..,n. 
\Bem{(Note: no connection is introduced between 
nodes contained  in $h_1$).
}         For k=1 proceed as if $h_1$ were a twig with an
empty set as a  branch for it. 
\AbbVier
 It is easily checked that 
the hypergraph induced by a belief network structure obtained in this way is 
in fact a hypertree (if reduced, then exactly the original reduced 
hypertree). Let us turn now to valuations (Fig.\ref{abbvier}.a).  
Let $Bel_i$ be the valuation originally attached to the hyperedge $h_i$. 
then $Bel = Bel_1 \oplus ...\oplus Bel_n$. 
What conditional belief is to be 
attached to $h_n$ ? First marginalize: $Bel'_n = Bel_1^{\downarrow h_1 \cap 
h_n} \oplus \dots \oplus Bel_{n-1}^{\downarrow h_{n-1} \cap 
h_n} \oplus Bel_n$. ( (Fig.\ref{abbvier}.b,.c))
Now calculate: $Bel"_n={Bel'}_n ^{|h_n \cap h_{i_n}}$, and 
$Bel"'_n={Bel'}_n  ^{\downarrow h_n \cap h_{i_n}}$. 
Let  $Bel_{*k}= Bel_k\ominus Bel_k ^{\downarrow h_1 \cap 
h_n}$  for k=1,...,$i_n$-1,$i_n$+1,...,(n-1),   (Fig.\ref{abbvier}.d) 
and 
let $Bel_{*i_n}= (Bel_{i_n}\ominus Bel_{i_n} ^{\downarrow h_1 \cap 
h_n}) \oplus Bel"'_n$ .
Obviously, $Bel=Bel_{*1} 
\oplus 
\dots \oplus Bel_{*(n-1)} \oplus Bel"_n$  (Fig.\ref{abbvier}.e).
Now let us consider a new hypertree only with hyperedges $h_1,\dots 
h_{n-1}$, and 
with valuations equal to those marked with asterisk (*), and repeat the 
process 
until only one hyperedge is left, the now valuation of which is considered 
 as $Bel"_1$. In the process, a new factorization is 
obtained: $Bel=Bel"_1 \oplus \dots \oplus  Bel"_n$. \\
If now for a hyperedge $h_k$ $card(h_k-h_{i_k})=1$, then we assign $Bel"_k$ 
to 
the node of the belief network corresponding to $h_k-h_{i_k}$. If  for a 
 hyperedge $h_k$ $card(h_k-h_{i_k})>1$, then we split  $Bel"_k$ as follows: 
Let $h_k-h_{i_k}=\{X_{k1},X_{k2},....,X_{km}\}$ and the indices shall 
correspond to the order in the belief network induced by the above 
construction procedure. Then 
$$Bel"_k=Bel ^{\downarrow h_k|h_k \cap h_{i_k}}=  
 \bigoplus_{j=1}^{m} Bel ^{\downarrow (h_k \cap h_{i_k}) \cup 
\{X_{k1},...,X_{kj}\} | (h_k \cap h_{i_k}) \cup 
\{X_{k1},...,X_{kj}\}-\{X_{kj}\}}$$
and we assign valuation $Bel ^{\downarrow (h_k \cap h_{i_k}) \cup 
\{X_{k1},...,X_{kj}\} | (h_k \cap h_{i_k}) \cup 
\{X_{k1},...,X_{kj}\}-\{X_{kj}\}}$ to the node corresponding to $X_{kj}$ in 
the network structure. It is easily checked that:
\begin{th} \label{xxxx}  \cite{Klopotek:93f}
(i) The network obtained by the above construction of its structure and 
valuation from hypertree factorization is a belief network.\\
(ii) This belief network represents exactly the joint belief distribution of 
the hypertree\\
(iii) This belief network induces exactly the original reduced hypertree 
structure
\end{th}
The above theorem implies that any hypergraph suitable for 
propagation must have a compatible 
belief network. Hence seeking for belief network decompositions of joint 
belief  distributions  is  sufficient  for  finding  any 
factorization suitable for Shenoy/Shafer propagation of uncertainty.  

\section{Discussion}

We presented here selected concepts of structuring of DS belief functions with
a special  emphasis on their capability to capture independence from the point
of view of the claim that belief functions generalize bayes notion
of probability.\\

 It is demonstrated that Zhu and Lee's \cite{Zhu:93} logical
networks and Smets' \cite{Smets:93} directed acyclic graphs are unable to
capture statistical dependence/independence of bayesian networks
\cite{Pearl:88}.
  Zhu and Lee's \cite{Zhu:93} just assume conditional
independence of premise and conclusion of an implication and hence assume 
that the dominant value of premise implies the dominant value of conclusion.
This is wrong from the logical point of view because  the implication
may be inverse in the data, but the presence of cases for which the rule 
is not applicable will distort the conclusions of expert system based on
rules like those proposed by   Zhu and Lee's \cite{Zhu:93}. 
Smets \cite{Smets:93} goes to other extreme and assumes that all variables but
in very restricted cases are dependent. His definition of cognitive
independence applies to bayesian belief functions only in those rare cases of
uniform joint probability distribution. Smets' notion of conditional
independence for bayesian belief functions means that the conditioning
variable is degenerate from statostical point of view: it takes only one value
and hence is usually ommitted from any statistical analysis. So conditional
independence again reduces to uniform distribution of conditioned variables. \\

On the other hand, though Shenoy and Shafer's hypergraphs
can explicitly represent bayesian network factorization of bayesian belief
functions, they disclaim any need for representation of independence of
variables in belief functions. Cano et al. \cite{Cano:93} reject the
hypergraph representation of Shenoy and Shafer just on  grounds of missing
 representation of variable independence, but in their frameworks some
belief functions factorizable in Shenoy/Shafer  framework  cannot 
be factored.
The approach in \cite{Klopotek:93f} on the other hand combines the merits of
both Cano et al. and of Shenoy/Shafer approach in that for Shenoy/Shafer
approach no simpler factorization than that in \cite{Klopotek:93f} approach
exists and on the other hand all independences among variables captured in
Cano et al. framework and many more are captured in \cite{Klopotek:93f}
approach.

Two new operators for MTE have been introduced in connection with the
approach of \cite{Klopotek:93f}: decombination $\ominus$ and anti-conditioning
$|$. We shall stress that operators in this sense have been also introduced
previously by Shenoy in \cite{Shenoy:91} in connection with his  valuation
networks. We shall emphasize however one important difference making our
approach more general. Shenoy  insists that belief functions suitable for
anti-conditioning should have unique unity item which means that the belief
function shall have a focal point equal to the universe. This excludes
bayesian belief functions so that Shenoy's framework could not represent many
of independences explicated within our framework. 

Inversion of Shafer's conditioning $Bel(.||B)$ was also considered by Cano et
al. \cite{Cano:93},
 Smets \cite{Smets:93} and Shafer \cite{Shafer:76}. The 'a priori
conditioning' of Cano et al. \cite{Cano:93} is a special case of our
anti-conditioning in that it requires marginalization on conditioning
variables to yield a vacuous belief function. On the other hand deconditioning
of Smets \cite{Smets:93} and conditional embedding of Shafer \cite{Shafer:76}
are entirely different in nature. Our approach
starts with joint absolute belief distribution and 
 tries to remove the impact of
absolute distribution of anti-conditioning variables from the picture of
relation
between the anti-conditioning and the anti-conditioned variable.
Deconditioning
and conditional embedding 
start with selected conditional beliefs 
try to establish a joint absolute belief
distribution of conditioned and conditioning variables. Hence the concept of
anti-conditioning is orthogonal to that of conditional embedding and
deconditioning.

\section{Conclusions}

For many approaches to structuring of belief functions the claim of
generalization of bayesian probabilities by Dempster-Shafer belief functions 
is actually illusive. Especially

\begin{itemize}
\item  Zhu and Lee's \cite{Zhu:93} logical
networks and Smets' \cite{Smets:93} directed acyclic graphs are unable to
capture statistical dependence/independence of bayesian networks
\item 
Though Shenoy and Shafer's hypergraphs can
explicitly represent bayesian network factorization of bayesian belief
functions, they disclaim any need for representation of independence of
variables in belief functions. 
\item 
In Cano et al. \cite{Cano:93}  frameworks some
belief functions factorizable in Shenoy/Shafer  framework  cannot 
be factored.
\item 
The approach in \cite{Klopotek:93f} on the other hand combines the merits of
both Cano et al. 
(representation of independences)
and of Shenoy/Shafer approach (Shenoy/Shafer
approach yields
 no simpler factorization than that in \cite{Klopotek:93f} approach)
\end{itemize}

\newcommand{\LitStelle}[2]{\bibitem{#1}}  
\newcommand{\A}[2]{#2 #1} 


\newcommand{\eng}[1]{}    
\newcommand{\pol}[1]{}    
\newcommand{\deu}[1]{}    

\renewcommand{\eng}[1]{#1}    

\newcommand{\IN}{\pol{[w:]}\deu{[in:]}\eng{[in:]}} 

\newcommand{\ReadingsIn}{G. Shafer, J. Pearl eds: 
{\it Readings in Uncertain 
Reasoning}, (ISBN 1-55860-125-2, 
Morgan Kaufmann Publishers Inc., San Mateo, California, 1990)}

\section*{Appendix: Basic Definitions of MTE}

\begin{df} 
Let $\Xi$ be a finite  set of elements called elementary events. 
Any subset of $\Xi$ be a composite event. $\Xi$ be called also the 
frame of discernment.\\
A basic probability assignment function m:$2^\Xi  \rightarrow [0,1]$
such that  $$  \sum_{A \in 2^\Xi } |m(A)|=1 $$
$$  m(\emptyset)=0 $$
$$\forall_{A \in 2^\Xi} \quad  0 \leq  \sum_{A \subseteq B} m(B)$$
($|.|$ - absolute value.\\
      
A belief function be defined as Bel:$2^\Xi \rightarrow [0,1]$ so that 
 $$Bel(A) = \sum_{B \subseteq A} m(B)$$
A plausibility function be Pl:$2^\Xi \rightarrow [ 0,1]$  with 
$$\forall_{A \in 2^\Xi} \  Pl(A) = 1-Bel(\Xi-A )$$
A commonalty function be Q:$2^\Xi \rightarrow [0,1]$ with 
 $$\forall_{A \in 2^\Xi} \quad Q(A) = \sum_{A \subseteq B} m(B)$$
\end{df}

For a belief function, any set A such that m(A) differs from zero, is called
{\em focal point}. A belief function where every focal point is a set
with cardinality 1 (singleton) is called {\em bayesian belief function}.\\

Furthermore, a Rule of Combination of two Independent Belief Functions 
$Bel_1$,
 $Bel_2$ Over the Same Frame of Discernment (the so-called Dempster-Rule),
denoted 
    $$Bel_{E_1,E_2}=Bel_{E_1} \oplus Bel_{E_2}$$ 
 is defined as follows: :
$$m_{E_1,E_2}(A)=c \cdot  \sum_{B,C; A= B \cap C} m_{E_1}(B) \cdot 
m_{E_2}(C)$$ (c - constant normalizing the sum of $|m|$ to 1)

Furthermore, let the frame of discernment $\Xi$ be structured in that it is
identical to cross product of domains $\Xi_1$, $\Xi_2$, \dots $\Xi_n$ of n
discrete variables $X_1, X_2, \dots X_n$, which span the space $\Xi$. Let
$(x_1, x_2, \dots x_n)$ be a vector in the space spanned by the variables 
$X_1,
 ,  X_2, \dots X_n$. Its projection onto the subspace spanned by variables 
$X_{j_1}, X_{j_2}, \dots X_{j_k}$ ($j_1, j_2,\dots j_k$ distinct indices from
the set 1,2,\dots,n) is then the vector $(x_{j_1}, x_{j_2}, \dots x_{j_k})$. 
$(x_1, x_2, \dots x_n)$ is also called an extension of $(x_{j_1}, x_{j_2},
\dots x_{j_k})$. A projection of a set $A$ of such vectors is the set
$A ^{\downarrow X_{j_1}, X_{j_2}, \dots X_{j_k}}$ 
 of
projections of all individual vectors from A onto $X_{j_1}, X_{j_2}, \dots
X_{j_k}$. A is also called an extension of $A ^{\downarrow X_{j_1}, X_{j_2},
\dots X_{j_k}}$. A is called the vacuous extension of $A ^{\downarrow
X_{j_1},
 X_{j_2}, \dots X_{j_k}}$  iff A contains all possible extensions of each
individual vector in $A ^{\downarrow X_{j_1}, X_{j_2}, \dots X_{j_k}}$ .
The fact, that A is a vacuous extension of B onto space $X_1,X_2,\dots\,
X_n$ is denoted by $A=B ^{\uparrow X_1,X_2,\dots\,X_n}$
\begin{df}
Let m be a basic probability assignment function on the space of discernment
spanned by variables   $X_1,X_2,\dots\,X_n$. $m ^{\downarrow X_{j_1},
X_{j_2}, \dots X_{j_k}}$ is  called  the  projection  of  m  onto 
subspace spanned by
$X_{j_1}, X_{j_2}, \dots X_{j_k}$ iff 
$$m ^{\downarrow X_{j_1}, X_{j_2}, \dots X_{j_k}}(B)= c \cdot
\sum_{A; B=A  ^{\downarrow X_{j_1}, X_{j_2}, \dots X_{j_k}} } m(A)$$
(c - normalizing factor)
\end{df}
\begin{df}
Let m be a basic probability assignment function on the space of discernment
spanned by variables  $  X_{j_1},
X_{j_2}, \dots X_{j_k} $. $m ^{\uparrow X_1,X_2,\dots\,X_n}$ is called
the vacuous extension 
 of m onto superspace spanned by $X_1,X_2,\dots\,X_n$
iff 
$$m ^{\uparrow X_1, X_2, \dots X_n}(B ^{\uparrow X_1,X_2,\dots\,X_n})=m(B)$$

and $m ^{\uparrow X_1, X_2, \dots X_n}(A)=0$ for any other A. \\
We say that a belief function is vacuous iff $m(\Xi)=1$ and $m(A)=0$ for any A
different from $\Xi$.
\end{df}

Projections and vacuous extensions of Bel, Pl and Q functions are defined
with
respect to operations on m function. Notice that by convention if we want to
combine by Dempster rule two belief functions not sharing the frame of
discernment, we look for the closest common vacuous extension of their
frames of discernment without explicitly notifying it.

\begin{df} (See \cite{Shafer:90b}) Let B be a subset of $\Xi$, called 
evidence,
 $m_B$ be a basic probability assignment such that $m_B(B)=1$ and $m_B(A)=0$
for any A different from B. Then the conditional belief function $Bel(.||B)$
representing the belief function $Bel$ conditioned on evidence  B 
is defined
as: $Bel(.||B)=Bel \oplus Bel_B$. 
\end{df}

Notice: Vacuous extension is also called cylindric extension. In Smets'
interpretation of Dempster-Shafer theory names bel, pl and q are used instead
of Bel, Pl and Q. The justification is that Smets allows for $\emptyset$ to be
a focal point (open world assumption). 

In this paper, the notion of pseudo-belief functions is also  used
(last section before discussion). 
 A
pseudo-belief
function differs from proper belief function in that m is allowed to take
also negative values, but only in such a way as to ensure that Q remains
non-negative. However, Bel and Pl may get negative for pseudo-belief
functions.

 \end{document}